\documentclass[11pt,a4paper]{article}
\usepackage[hyperref]{acl2019}
\usepackage{times}
\usepackage{latexsym}

\usepackage{CJK}
\usepackage{arydshln}
\usepackage{multirow}
\usepackage{booktabs}
\usepackage{amsmath} 
\usepackage{amsfonts}
\usepackage{graphicx} 
\usepackage{subfigure}
\usepackage{amstext}
\usepackage{tikz}
\usepackage{pgfplots}

\usepackage{url}

\aclfinalcopy 
\def\aclpaperid{330} 

\title{ReCoSa: Detecting the Relevant Contexts with Self-Attention for Multi-turn Dialogue Generation}

\author{Hainan Zhang, Yanyan Lan, Liang Pang, Jiafeng Guo and Xueqi Cheng \\
CAS Key Lab of Network Data Science and Technology,\\ Institute of Computing Technology, Chinese Academy of Sciences\\
University of Chinese Academy of Sciences, Beijing, China\\
 \{zhanghainan,lanyanyan, pangliang, guojiafeng, cxq\}@ict.ac.cn\\
}

\date{}

\begin{document}
\maketitle
\begin{abstract}

In multi-turn dialogue generation, response is usually related with only a few contexts. Therefore, an ideal model should be able to detect these relevant contexts and produce a suitable response accordingly. However, the widely used hierarchical recurrent encoder-decoder models just treat all the contexts indiscriminately, which may hurt the following response generation process. Some researchers try to use the cosine similarity or the traditional attention mechanism to find the relevant contexts, but they suffer from either insufficient relevance assumption or position bias problem. In this paper, we propose a new model, named ReCoSa, to tackle this problem. Firstly, a word level LSTM encoder is conducted to obtain the initial representation of each context. Then, the self-attention mechanism is utilized to update both the context and masked response representation. Finally, the attention weights between each context and response representations are computed and used in the further decoding process. Experimental results on both Chinese customer services dataset and English Ubuntu dialogue dataset show that ReCoSa significantly outperforms baseline models, in terms of both metric-based and human evaluations. Further analysis on attention shows that the detected relevant contexts by ReCoSa are highly coherent with human\rq{}s understanding, validating the correctness and interpretability of ReCoSa.

\end{abstract}
\section{Introduction}
\begin{CJK*}{UTF8}{gbsn}

This paper is concerned with the multi-turn dialogue generation task, which is critical in many natural language processing (NLP) applications, such as customer services, intelligent assistant and chatbot. Recently, the hierarchical recurrent encoder-decoder (HRED) models~\cite{SERBAN:HRED1,SORDONI:HRED2} have been widely used in this area. In the encoding phase of these HRED models, a recurrent neural network (RNN) based encoder is first utilized to encode each input a context to a vector, and then a hierarchical RNN is conducted to encode these vectors to one vector. In the decoding phase, another RNN decoder is used to generate the response based on the above vector. The parameters of both encoder and decoder are learned by maximizing the averaged likelihood of the training data. 

\begin{table}
\centering
\scriptsize
\newcommand{\tabincell}[2]{\begin{tabular}{@{}#1@{}}#2\end{tabular}}
\begin{tabular}{ll} 
\toprule
\multicolumn{2}{c}{The first example}\\
\toprule
context1 &  \tabincell{l}{你好，在吗？(Hello)}\\ 
context2 &  \tabincell{l}{有什么问题我可以帮您呢?（What can I do for you?）}\\ 
context3 &  \tabincell{l}{\textcolor{red}{保真吗？（Is this product fidelity?）}}\\ 
response  &  \tabincell{l}{我们的商品都是海外采购的~ 绝对保证是正品的}\\ 
                &  \tabincell{l}{(Our products are all purchased overseas~ Absolutely \\guaranteed to be genuine)}\\ 
\bottomrule
\multicolumn{2}{c}{The second example}\\
\toprule
context1 &  \tabincell{l}{\textcolor{red}{我有个交易纠纷，麻烦你看看有进度吗}}\\ 
              &  \tabincell{l}{ \textcolor{red}{(I have a trading dispute. Could you please tell me} \\ \textcolor{red}{whether it is progressing? )}}\\ 
context2 &  \tabincell{l}{您好，请问是这个订单吗?(Hello, is this order?)}\\ 
context3 &  \tabincell{l}{对(Yes)}\\ 
response  &  \tabincell{l}{等待纠纷处理(Waiting for dispute resolution)}\\
\bottomrule
\end{tabular}
\caption{The two examples from the customer services dataset, and the red sentence indicates the relevant context to the response.} \label{tb:examples}
\end{table}

However, for this task, it is clear that the response is usually dependent on some relevant contexts, rather than all the context information. Here we give two examples, as shown in Table~\ref{tb:examples}. In the first example, the response is clearly related to the closest context, i.e.~post, in the first example. While in the second example, the response is related to context1. In these cases, if we use all contexts indiscriminately, as in HRED, it is likely that many noises will be introduced to the model, and the generation performance will be hurt significantly. Therefore, it is critical to detect and use the relevant contexts for multi-turn dialogue generation.

To tackle this problem, some researchers try to define the relevance of a context by using the similarity measure, such as the cosine similarity in Tian et al.~\cite{YAN:HARD}. However, the cosine similarity is conducted between each context and the post, with the assumption that the relevance between a context and a response is equivalent to the relevance between the context and the corresponding post, which is clearly insufficient in many cases, e.g.~example 2 in Figure~\ref{fig:examples}. Some other researchers, e.g.~Xing et al.~\cite{XING:SOFT} make an attempt by introducing the traditional attention mechanism to HRED. However, some related contexts are far from the response in the multi-turn dialogue generation task, and the RNN-based attention model may not perform well because it usually biases to the close contexts~\cite{LONG:2001}, namely position bias problem. Therefore, how to effectively detect and use the relevant contexts remains a challenging problem in multi-turn dialogue generation.

In this paper, we propose a new model, namely ReCoSa, to tackle this problem. The core idea is to use the self-attention mechanism to measure the relevance between the response and each context. The motivation comes from the fact that self-attention is superior in capturing long distant dependency, as shown in~\cite{Vaswani:2017}. Specifically, we first use a word-level LSTM encoder to obtain the fixed-dimensional representation of each context. Then, we use the self-attention mechanism to get the context and masked response representations. Finally, we calculate the attention weight between the context and response representations as the relevant score, and conduct a decoder based on the related contexts to generate the corresponding response.

In our experiments, we use two public datasets to evaluate our proposed models, i.e.~Chinese customer services and English Ubuntu dialogue corpus. The results show that ReCoSa has the ability to produce more diverse and suitable responses than traditional HRED models and its attention variants. Besides, we conduct an analysis on attention, and the results show that the ReCoSa obtains higher coherence with the human labels, which indicate that the detected relevant contexts by our model are reasonable.

\section{Related Work}
Despite many existing research works on single-turn dialogue generation~\cite{LI:GAN,MOU:BACKFORWARD,ZHANG:COHER,ZHANG:CVAR}, multi-turn dialogue generation has gain increasing attention. One reason is that it is more accordant with the real application scenario, such as chatbot and customer services. More importantly, the generation process is more difficult since there are more context information and constrains to consider~\cite{CHEN:MVHRED,ZHANG:2018,ZHANG:IR,WU:IR,ZHOU:IR}, which poses great challenges for researchers in this area.

Serban et al.~\cite{SERBAN:HRED1} proposed HRED which uses the hierarchical encoder-decoder framework to model all the context sentences. Since then, the HRED based models have been widely used in different multi-turn dialogue generation tasks, and many invariants have been proposed. For example, Serban et al.~\cite{SERBAN:VHERD,SERBAN:MRRNN} proposed Variable HRED (VHRED) and MrRNN which introduce the latent variables into the middle state to improve the diversity of generated responses. 

However, simply treating all contexts indiscriminately is not proper for the application of multi-turn dialogue generation, since the response is only usually related to a few previous contexts. Therefore some researchers try to define the relevance of the context by the similarity measure. For example, Tian et al.~\cite{YAN:HARD} proposed a weighted sequence (WSeq) attention model for HRED, using the cosine similarity to measure the degree of the relevance. Specifically, they first calculate the cosine similarity between the post embedding and each context sentence embedding, and then use this normalized similarity score as the attention weight. We can see that their results are based on an assumption that the relevance between a context and a response is equivalent to the relevance between the context and the corresponding post. However, in many cases, this assumption is actually not proper. Recently, Xing et al.~\cite{XING:SOFT} has introduced the traditional attention model to HRED, and a new hierarchical recurrent attention network (HRAN) has been proposed, which is similar with the Seq2Seq model with attention~\cite{BAHDANAU:ATTENTION}. In this model, the attention weight is computed based on the current state, the sentence-level representation and the word-level representation. However, some relevant contexts in multi-turn dialogue generation are relatively far from the response, therefore the RNN-based attention model may not perform well because it usually biases to the close contexts~\cite{LONG:2001}. Shen et al.~\cite{CHEN:MVHRED} introduced the memory network into the VHRED model, so that the model can remember the context information. Theoretically, it can retrieve some relevant information from the memory in the decoding phase, however, it is not clearly whether and how the system accurately extracts the relevant contexts.

The motivation of this paper is how to effectively extract and use the relevant contexts for multi-turn dialogue generation. Different from previous studies, our proposed model can focus on the relevant contexts, with both long and short distant dependency relations, by using the self-attention mechanism.
\section{Relevant Context Self-Attention Model}
\begin{figure}
\centering
\includegraphics[width=0.45\textwidth]{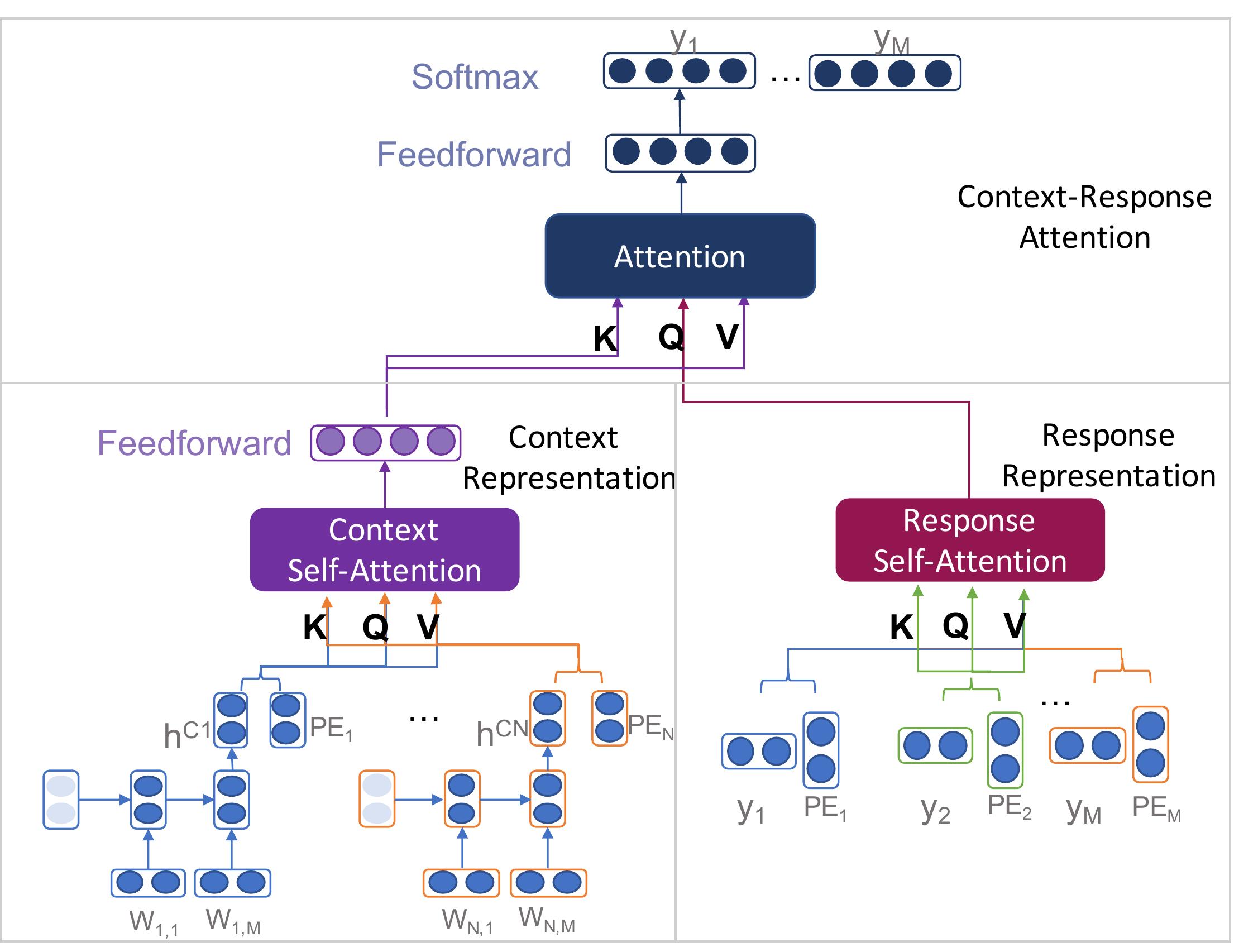}
\caption{The architecture of ReCoSa model} \label{fig:architecture}
\end{figure}

In this section, we will describe our relevant context with self-attention (ReCoSa) model in detail, with architecture shown in Figure~\ref{fig:architecture}. 
ReCoSa consists of a context representation encoder, a response representation encoder and a context-response attention decoder. For each part, we use the multi-head self-attention module to obtain the context representation, response representation and the context-response attention weights. Firstly, the word-level encoder encodes each context as a low-dimension representation. And then, a multi-head self-attention component transforms these representations and position embeddings to the context attention representation. Secondly, another multi-head self-attention component transforms the masked response\rq{}s word embedding and position embedding to the response attention representation. Thirdly, the third multi-head attention component feeds the context representation as key and value, and the response representation as query in the context-response attention module. Finally, a softmax layer uses the output of the third multi-head attention component to obtain the word probability for the generation process.

\subsection{Context Representation Encoder}
We will introduce the main components of the context representation encoder in this section. The word-level encoder first encodes each context as a fixed vector. And then the context self-attention module transforms each sentence vector to a context representation.
\subsubsection{Word-level Encoder}
We first introduce the LSTM-based word level encoder~\cite{BAHDANAU:ATTENTION} used in our model. 
Given the context set $C=\{s_1,\dots,s_N\}$, each sentence in $C$ is defined as $s_i=\{x_1,\dots,x_M\}$. Please note that in our paper the post is treated as the last context sentence $s_N$. Given a sentence $s_i$ as the input, a standard LSTM first encodes each input context to a fixed-dimension vector $h_M$ as follows.
\begin{equation*}
\begin{aligned}
i_k&=\sigma(W_i[h_{k-1},w_k]),\,\,f_k=\sigma(W_f[h_{k-1},w_k]),\\
o_k&=\sigma(W_o[h_{k-1},w_k]),\,\,l_k=\tanh(W_l[h_{k-1},w_k]),\\
c_k&=f_kc_{k-1}+i_kl_k,\,\,\,\,\,\,\,\,\,h_i=o_k\tanh(c_k),
\end{aligned}
\end{equation*}
where $i_k,f_k$ and $o_k$ are the input, memory and output gate, respectively. $w_k$ is the word embedding for $x_k$, and $h_k$ stands for the vector computed by LSTM at time $k$ by combining $w_k$ and $h_{k-1}$. $c_k$ is the cell at time $k$, and $\sigma$ denotes the sigmoid function. $W_i, W_f, W_o$ and $W_l$ are parameters. 
We use the vector $h_M$ as the sentence representation. Therefore, we obtain the sentence representations $\{h^{s_1},\dots,h^{s_N}\}$.

It has been widely accepted that the self-attention mechanism itself cannot distinguish between different positions. So it is crucial to encode each position information. Actually, there are various ways to encode positions, and the simplest one is to use an additional position embedding. In our work, we parameterized position embeddings $P_i \in \mathbb{R}^{d}, i=1,\dots,N$. The position embeddings are simply concatenated to the sentence representations. Finally, we obtain the sentences representation$\{(h^{s_1},P_1),\dots,(h^{s_N},P_N)\}$.

\subsubsection{Context Self-Attention}
Self-attention is a special attention mechanism to compute a sequence\rq{}s representation using only the sequence itself, which has been successfully applied to many tasks, such as machine translation, reading comprehension, summarization, and language understanding~\cite{Vaswani:2017,CHENG:2016,PARIKH:2016,paulus2017deep,shen2018disan}. One critical advantage of self-attention is that it has the ability to well capture the long distant dependency information~\cite{Vaswani:2017}. That\rq{}s why we use this mechanism in our work.

In this paper, we adopt the multi-head attention~\cite{Vaswani:2017} mechanism. Given a matrix of $n$ query vectors $Q \in \mathbb{R}^{n\times d}$, keys $K \in \mathbb{R}^{n\times d}$ and values $V \in \mathbb{R}^{n\times d}$, the scaled dot-product attention computes the
attention scores based on the following equation:
\begin{equation*}
Attention(Q,K,V)=softmax(\frac{QK^{T}}{\sqrt{d}})V,
\end{equation*}
where $d$ is the number of the hidden units in our network.

The $H$ parallel heads are used to focus on different parts of channels of the value vectors. Formally, for the i-th head, we denote the learned linear maps by $W_i^Q \in \mathbb{R}^{n\times d/H}$,$W_i^K \in \mathbb{R}^{n\times d/H}$ and $W_i^V \in \mathbb{R}^{n\times d/H}$, which correspond to queries, keys, and values, respectively. Then the scaled dot-product attention is used to calculate the relevance score between queries and keys, to output mixed representations. The mathematical formulation is:
\begin{equation*}
M_i=Attention(QW_i^Q,KW_i^K,VW_i^V).
\end{equation*}
Finally, all the vectors produced by parallel heads are concatenated together to form a single vector. Again, a linear map is used to mix different channels from different heads:
\begin{equation}\label{eq:oc}
\begin{aligned}
M&=Concat(M_1,\dots,M_H), \\
O &=MW,
\end{aligned}
\end{equation}
where $M \in \mathbb{R}^{n\times d}$ and $W \in \mathbb{R}^{d \times d}$.

To obtain the context representation, the multi-head attention mechanism first feeds the matrix of sentences representation vectors $\{(h^{s_1},P_1),\dots,(h^{s_N},P_N)\}$. as queries, keys and values matrices by using different linear projections. Then the context representation is computed as $O_s$ in equation~\ref{eq:oc}. 
We use a feedforward network to output the context attention representation $O_s^f$.

\subsection{Response Representation Encoder}\label{sec:response}
Given the response $Y=\{y_1,\cdots, y_M\}$ as the input, another multi-head self-attention component transforms each word embedding and its position embedding to obtain the response representation. For each word $y_t$, this multi-head attention component feeds the matrix of response vectors $\{(w_1+P_1),\cdots,(w_{t-1},P_{t-1})\}$ as queries, keys and values matrices by using different linear projections. Then the response\rq{}s hidden representation is computed as $O_r$ in equation~\ref{eq:oc}. 

After that, we use the mask operator on the response for the training. For each word $y_t$, we mask $\{y_{t+1},\cdots, y_M\}$ and only see $\{y_1,\cdots,y_{t-1}\}$. For inference, we use the loop function on the generated response $G$. Take the $t^{th}$ generation as an example. Given the context $C=\{s_1,\dots,s_N\}$ and the generated response $\{g_1,\cdots, g_{t-1}\}$, we feed $\{g_1,\cdots, g_{t-1}\}$ as the response representation to obtain the $t^{th}$ word distribution in the generation response.

\subsection{Context-Response Attention Decoder}
The third multi-head attention component feeds the context attention representation~$O_s^f$ as key and value, and the response hidden representation $O_r$ as query. The output is denoted as $O_d$. We also use a new feedforward network to obtain the hidden vector $O_d^f$, as conducted in section~3.1.2.

Finally, a softmax layer is utilized to obtain the word probability for the generation process.
Formally, given an input context sequences $C=\{s_1,\dots,s_N\}$, the log-likelihood of the corresponding response sequence $Y=\{y_1,\cdots, y_M\}$ is:
\begin{equation*}
log P(Y|C;\theta) = \sum_{t=1}^{M} log P(y_t|C,y_1,\cdots,y_{t-1};\theta).
\end{equation*}
Our model predicts the word $y_t$ based on the hidden representation $O_d^f$ produced by the topmost softmax layer:
\begin{equation*}
\begin{aligned}
P(y_t|C,y_1,\cdots,y_{t-1}; \theta) &= P(y_t|O_s^f; \theta) \\
                  &=softmax(W_o O_s^f),
\end{aligned}
\end{equation*}
where $W_o$ is the parameters. Our training objective is to maximize the log likelihood of the ground-truth words given the input contexts over the entire training set. Adam is used for optimization in our experiments.

\section{Experiments}
In this section, we conduct experiments on both Chinese customer service and English Ubuntu dialogue datasets to evaluate our proposed method.
\subsection{Experimental Settings}
We first introduce some empirical settings, including datasets, baseline methods, parameter settings, and evaluation measures.
\subsubsection{Datasets}
We use two public multi-turn dialogue datasets in our experiments. 
The Chinese customer service dataset, named JDC, consists of 515,686 conversational context-response pairs published by the JD contest\footnote{https://www.jddc.jd.com}. We randomly split the data to training, validation, and testing sets, which contains 500,000, 7,843 and 7,843 pairs, respectively.
The English Ubuntu dialogue corpus\footnote{https://github.com/rkadlec/ubuntu-ranking-dataset-creator} is extracted from the Ubuntu question-answering forum, named Ubuntu~\cite{UBUNTU:2015}. The original training data consists of 7 million conversations from 2004 to April 27,2012. The validation data are conversational pairs from April 27,2012 to August 7,2012, and the test data are from August 7,2012 to December 1,2012. We use the official script to tokenize, stem and lemmatize, and the duplicates and sentences with length less than 5 or longer than 50 are removed. Finally, we obtain 3,980,000, 10,000 and 10,000 pairs for training, validation and testing, respectively. 

\subsubsection{Baselines and Parameters Setting}
Six baseline methods are used for comparison, including traditional Seq2Seq~\cite{SUTSKEVER:S2S}, HRED~\cite{SERBAN:HRED1}, VHRED~\cite{SERBAN:VHERD}, Weighted Sequence with Concat (WSeq)~\cite{YAN:HARD}, Hierarchical Recurrent Attention Network (HRAN)~\cite{XING:SOFT} and Hierarchical Variational Memory Network (HVMN)~\cite{CHEN:MVHRED}. 

For JDC, we utilize the Chinese word as input. Specifically, we use the Jieba tool for word segmentation, and set the vocabulary size as 69,644. For Ubuntu, the word vocabulary size is set as 15,000. For a fair comparison among all the baseline methods and our methods, the numbers of hidden nodes are all set to 512, and batch sizes are set to 32. The max length of dialogue turns is 15 and the max sentence length is 50. The head number of ReCoSa model is set as 6. Adam is utilized for optimization, and the learning rate is set to be 0.0001. We run all the models on a Tesla K80 GPU card with Tensorflow\footnote{https://github.com/zhanghainan/ReCoSa}.

\subsubsection{Evaluation Measures}
We use both quantitative metrics and human judgements for evaluation in our experiment. Specifically, we use two kinds of metrics for quantitative comparisons. One kind is traditional metrics, such as PPL and BLEU score~\cite{XING:TOPIC}, to evaluate the quality of generated responses. They are both widely used in NLP and multi-turn dialogue generation~\cite{CHEN:MVHRED,YAN:HARD,XING:SOFT}. The other kind is the recently proposed {\em distinct}~\cite{LI:DRL}, to evaluate the degree of diversity of the generated responses by calculating the number of distinct unigrams and bigrams in the generated responses.

\begin{table}
\centering
\scriptsize
\begin{tabular}{lcccc}
\multicolumn{5}{c}{JDC Dataset}\\
\toprule
model & PPL & BLEU &  distinct-1 & distinct-2 \\ \hline
SEQ2SEQ & 20.287 & 11.458 & 1.069 & 3.587 \\
HRED & 21.264 & 12.087 & 1.101 & 3.809 \\
VHRED & 22.287 & 11.501 & 1.174 & 3.695 \\ \hline
WSeq & 21.824 & 12.529 & 1.042 & 3.917 \\
HRAN & 20.573 & 12.278 & \bf{1.313} & 5.753 \\
HVMN & 22.242 & 13.125 & 0.878 & 3.993  \\ \hline
ReCoSa & \bf{17.282} & \bf{13.797} & 1.135 & \bf{6.590} \\
\bottomrule
\multicolumn{5}{c}{Ubuntu Dataset}\\
\toprule
model &PPL & BLEU & distinct-1 & disttinct-2 \\ \hline
SEQ2SEQ & 104.899 & 0.4245 & 0.808 & 1.120 \\
HRED & 115.008 & 0.6051 & 1.045 & 2.724\\
VHRED & 186.793 & 0.5229 & 1.342 & 2.887\\ \hline
WSeq & 141.599 & 0.9074 & 1.024 & 2.878\\
HRAN & 110.278 & 0.6117 & 1.399 & 3.075\\
HVMN & 164.022 & 0.7549 & 1.607 & 3.245\\ \hline
ReCoSa & \bf{96.057} & \bf{1.6485} & \bf{1.718} & \bf{3.768}\\
\bottomrule
\end{tabular}
\caption{\label{tb:metricJDC}  The metric-based evaluation results (\%). }
\end{table}

\begin{table*}
\scriptsize
\centering
\begin{tabular}{lccccccccccccccc}
\multicolumn{15}{c}{JDC Dataset}\\
\toprule
model & P@1 & R@1 & F1@1&  P@3 & R@3 &F1@3& P@5 & R@5 &F1@5&  P@10 & R@10&F1@10 \\ \hline
WSeq & \bf{35.20} & \bf{29.73} & \bf{16.12} & 24.27 & 51.49 & 16.50 & 21.61 & 71.76 & 16.61 & 17.45 & 97.17 & 14.79 \\ 
HRAN & 22.88 & 15.56 & 9.26 & 24.13 & 46.22 & 15.85 & 22.78 & 66.22 & 16.95 & 21.05 & 91.11 & 17.10  \\ \hline
ReCoSa-head1 & 25.98 & 19.19 & 11.04 & 25.35 & 52.33 & 17.08 & 23.92 & 73.84 & 18.07 & \bf{22.55} & \bf{97.67} & \bf{18.32}\\
ReCoSa-head2 & 17.32 & 12.79 & 7.36 & 24.23 & 50.00 & 16.32 & 24.29 & 75.00 & 18.35 & 22.15 & 95.93 & 17.99 \\
ReCoSa-head3 & 27.56 & 20.35 & 11.71 & \bf{26.20} & \bf{54.07} & \bf{17.65} & 23.92 & 73.84 & 18.07 & 22.01 & 95.35 & 17.88 \\
ReCoSa-head4 & 20.47 & 15.12 & 8.70 & 25.92 & 53.49 & 17.46 & 23.92 & 73.84 & 18.07 & \bf{22.55} & \bf{97.67} & \bf{18.32} \\
ReCoSa-head5 & 29.92 & 22.09 & 12.71 & 25.92 & 53.49 & 17.46 & \bf{24.67} & \bf{76.16} & \bf{18.63} & 22.15 & 95.93 & 17.99 \\
ReCoSa-head6 & 25.20 & 18.60 & 10.70 & 25.35 & 52.33 & 17.08 & 24.29 & 75.00 & 18.35 & 22.15 & 95.93 & 17.99 \\
\bottomrule
\end{tabular}
\caption{\label{tb:ATTENTION}  The attention analysis results (\%). }
\end{table*}

\begin{table}[t]
\centering
\small
\begin{tabular}{lcccc} 
\multicolumn{5}{c}{JDC Dataset}\\
\toprule
\multirow{2}{1cm}{model} & \multicolumn{3}{c}{ReCoSa vs.}  & \multirow{2}{0.5cm}{kappa}\\ \cline{2-4}
 & win (\%) & loss (\%) & tie (\%)\\ \hline
SEQ2SEQ & 53.45 & 3.45 & 43.10&0.398\\
HRED & 44.83 & 10.34 & 44.83&0.373\\
VHRED & 50.00 & 6.90 & 43.10&0.369\\ \hline
WSeq & 34.48 & 8.62 & 56.90&0.379\\
HRAN & 24.14 & 13.79 & 62.07&0.384\\
HVMN & 27.59 & 13.79 & 58.62&0.383\\
\bottomrule
\multicolumn{5}{c}{Ubuntu Dataset}\\
 \toprule
\multirow{2}{1cm}{model} & \multicolumn{3}{c}{ReCoSa vs.}  & \multirow{2}{0.5cm}{kappa}\\ \cline{2-4}
 & win (\%) & loss (\%) & tie (\%)\\ \hline
SEQ2SEQ & 55.32 & 2.13 & 42.55&0.445\\
HRED & 44.68 & 8.51 & 46.81&0.429\\
VHRED & 48.94 & 8.51 & 42.55&0.421\\ \hline
WSeq & 25.53 & 14.89 & 59.57&0.440\\
HRAN & 34.04 & 10.64 & 55.32&0.437\\
HVMN & 27.66 & 12.77 & 59.57&0.434\\
\bottomrule
\end{tabular}
\caption{\label{tb:humanEvaluation}  The human evaluation on JDC and Ubuntu.}
\end{table}

For human evaluation, given 300 randomly sampled context and their generated responses, three annotators (all CS majored students) are required to give the comparison between ReCoSa model and baselines, e.g. win, loss and tie, based on the coherence of the generated response with respect to the contexts. For example, the win label means that the generated response of ReCoSa is more proper than the baseline model.
\subsection{Experimental Results}
Now we demonstrate our experimental results on the two public datasets.

\subsubsection{Metric-based Evaluation}
The quantitative evaluation results are shown in Table~\ref{tb:metricJDC}. From the results, we can see that the attention-based models, such as WSeq, HRAN and HVMN, outperform the traditional HRED baselines in terms of BLEU and {\em distinct-2} measures. That\rq{}s because all these models further consider the relevance of the contexts in the optimization process. HRAN uses a traditional attention mechanism to learn the importance of the context sentences. HVMN uses a memory network to remember the relevant context. But their effects are both quite limited. Our proposed ReCoSa performs the best. Take the BLEU score on JDC dataset for example, the BLEU score of ReCoSa model is 13.797, which is significantly better than that of HRAN and HVMN, i.e.,~12.278 and 13.125. The {\em distinct} scores of our model are also higher than baseline models, which indicate that our model can generate more diverse responses. We have conducted the significant test, and the result shows that the improvements of our model are significant on both Chinese and English datasets, i.e., $\text{p-value}<0.01$. In summary, our ReCoSa model has the ability to produce high quality and diverse responses, as compared with baseline methods.
\subsubsection{Human Evaluation}
The human evaluation results are shown in Table~\ref{tb:humanEvaluation}. The percentage of win, loss and tie, as compared with the baselines, are given to evaluate the quality of generated responses by ReCoSa. From the results, we can see that the percentage of win is always larger than that of loss, which shows that our ReCoSa model significantly outperforms baselines. Take JDC as an example. Compared with HRAN, WSeq and HVMN, the ReCoSa achieves preference gains (win subtracts loss) 10.35\%, 25.86\% and 13.8\%, respectively. Kappa~\cite{KAPPA:1971} value is presented to demonstrate the consistency of different annotators. 
We also conducted the significant test, and the result shows that the improvements of our model are significant on both two datasets, i.e., $\text{p-value}<0.01$.

\begin{table*}
\centering
\scriptsize
\newcommand{\tabincell}[2]{\begin{tabular}{@{}#1@{}}#2\end{tabular}}
\begin{tabular}{lll} 
\multicolumn{3}{c}{Example1}\\
\toprule
\textcolor{red}{ context1} &  \tabincell{l}{ \textcolor{red}{ 货物没有收到但是申请了退货}} & \tabincell{l}{  \textcolor{red}{ The goods were not received, but the return was applied}}\\ 
context2 &  \tabincell{l}{请您稍等,马上为您查询~}& \tabincell{l}{ Please wait, I will check for you right away.}\\ 
\textcolor{red}{ context3} &  \tabincell{l}{\textcolor{red}{ 这个申请什么时候能通过?}}& \tabincell{l}{ \textcolor{red}{ When will this application pass?}}\\ 
{response}  &  \tabincell{l}{{ 这个订单已经发货了,需要到了站点之后处理}}& \tabincell{l}{{  This order has been shipped and should be processed after shipping to the site.}}\\  \hline
Seq2Seq &  \tabincell{l}{您好,请问有什么问题可以帮您?}&\tabincell{l}{ What can I do for you?}\\ 
HRED &  \tabincell{l}{亲爱的客户,还麻烦您提供下订单号,妹子这边给您查询哦~}& \tabincell{l}{ Please give me your order ID. I will check for you now}\\ 
VHRED &  \tabincell{l}{您好,很高兴为您服务}& \tabincell{l}{I am very happy to serve you.}\\  \hline
WSeq &  \tabincell{l}{申请中}& \tabincell{l}{ Applying}\\ 
HRAN &  \tabincell{l}{是已经完成了哦,系统无法拦截站点}& \tabincell{l}{It\rq{}s already done, and the system can\rq{}t intercept the site.}\\ 
HVMN &  \tabincell{l}{建议等待呢}& \tabincell{l}{ Please wait}\\  \hline
{ReCoSa} &  \tabincell{l}{  亲爱的,查看到您的服务单还没有更新,请您耐心等待一下}& \tabincell{l}{   Dear, check that your service order has not been updated yet. Please wait.}\\ 
\bottomrule
\end{tabular}
\caption{\label{tb:caseJDC1}  The generated response Example1 from different models on JDC. The red contexts indicate the relevant context to the response.}
\end{table*}

\begin{figure}[!t]
\scriptsize
\centering
\subfigure[\scriptsize ReCoSa-head1.]{
\begin{minipage}[t]{0.4\linewidth}
\centering
\includegraphics[width=1.4in]{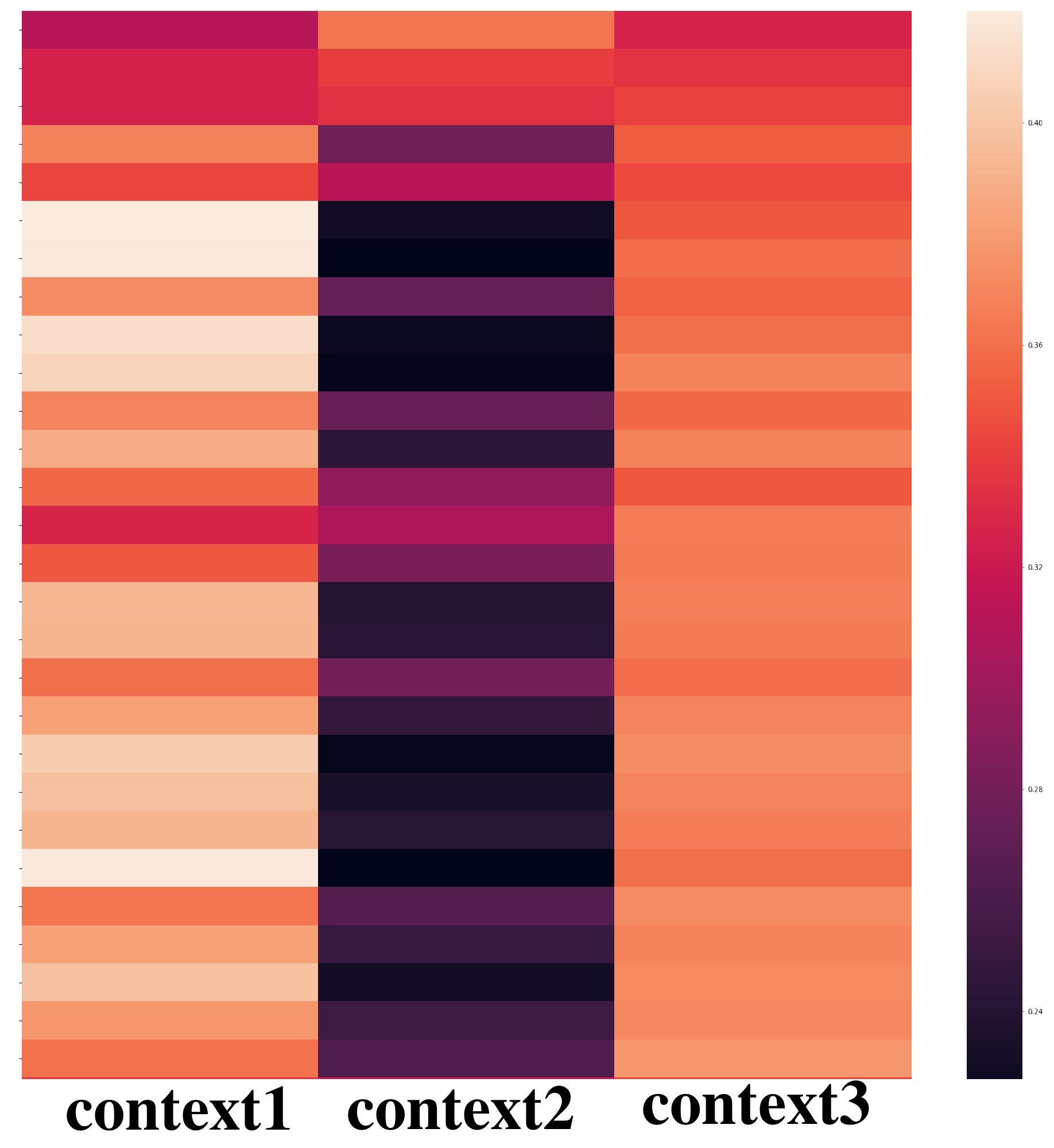}
\end{minipage}%
}%
\subfigure[\scriptsize ReCoSa-head2.]{
\begin{minipage}[t]{0.4\linewidth}
\centering
\includegraphics[width=1.4in]{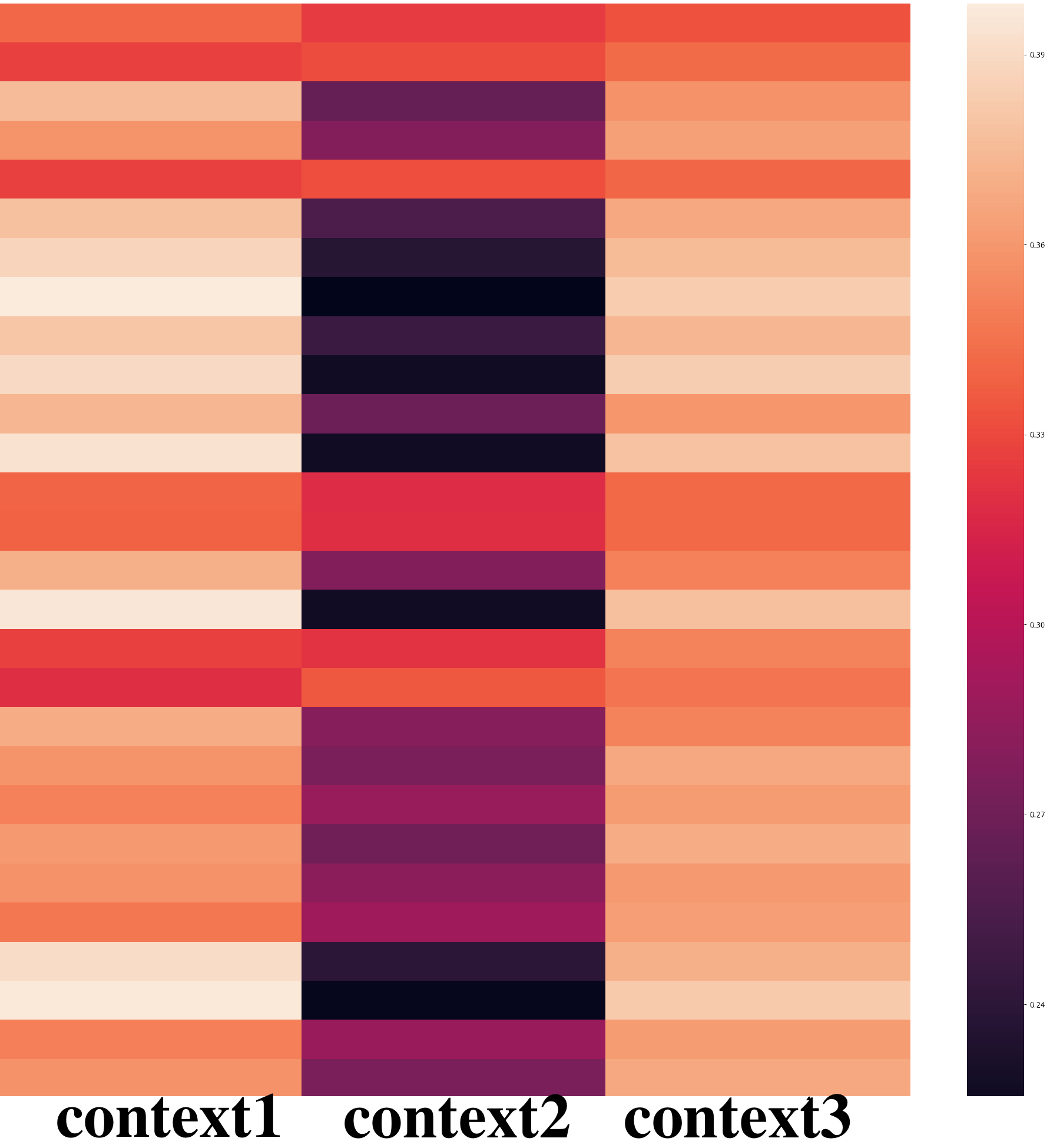}
\end{minipage}%
}%

\subfigure[\scriptsize ReCoSa-head3.]{
\begin{minipage}[t]{0.4\linewidth}
\centering
\includegraphics[width=1.4in]{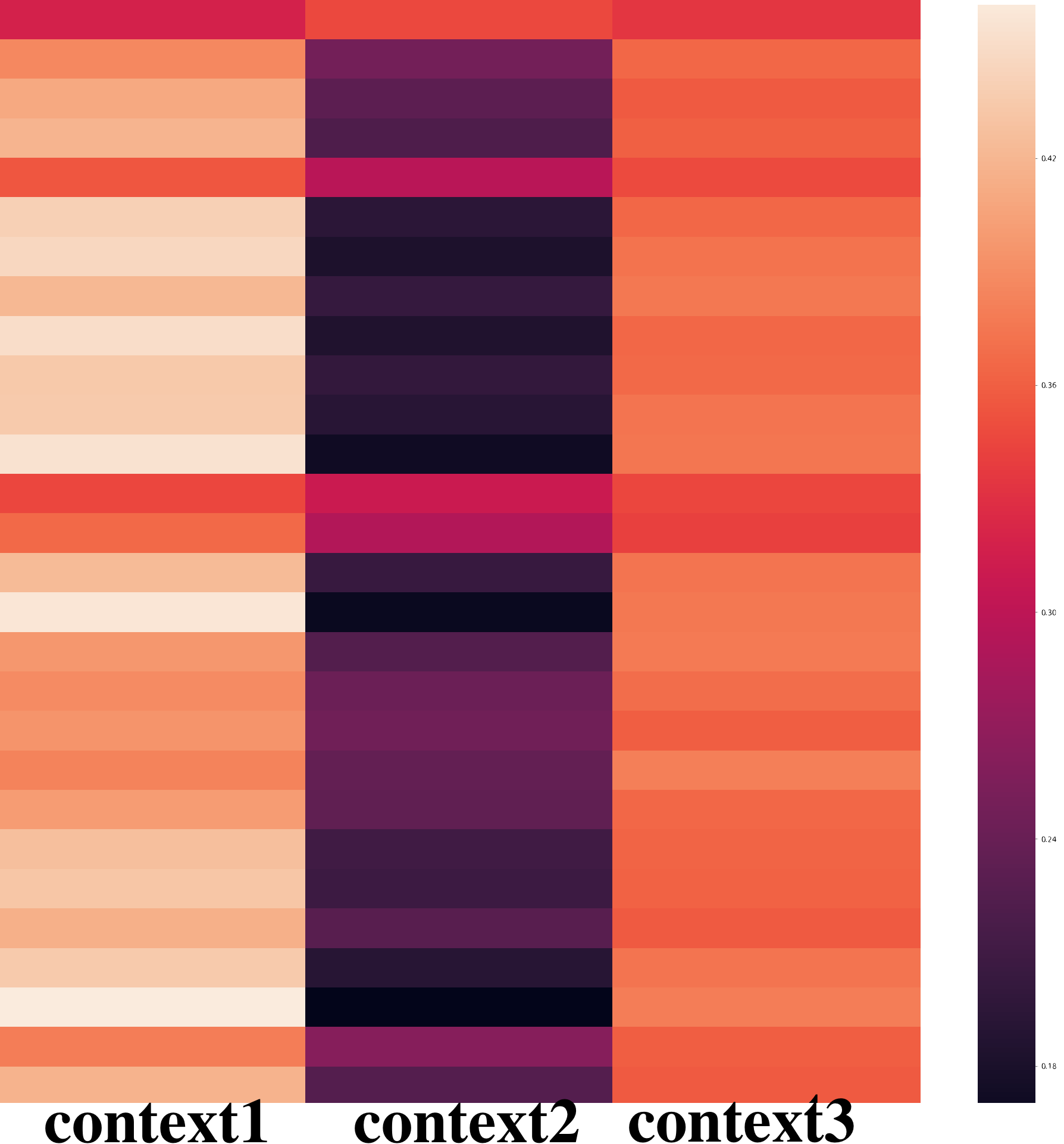}
\end{minipage}
}%
\subfigure[\scriptsize ReCoSa-head4.]{
\begin{minipage}[t]{0.4\linewidth}
\centering
\includegraphics[width=1.4in]{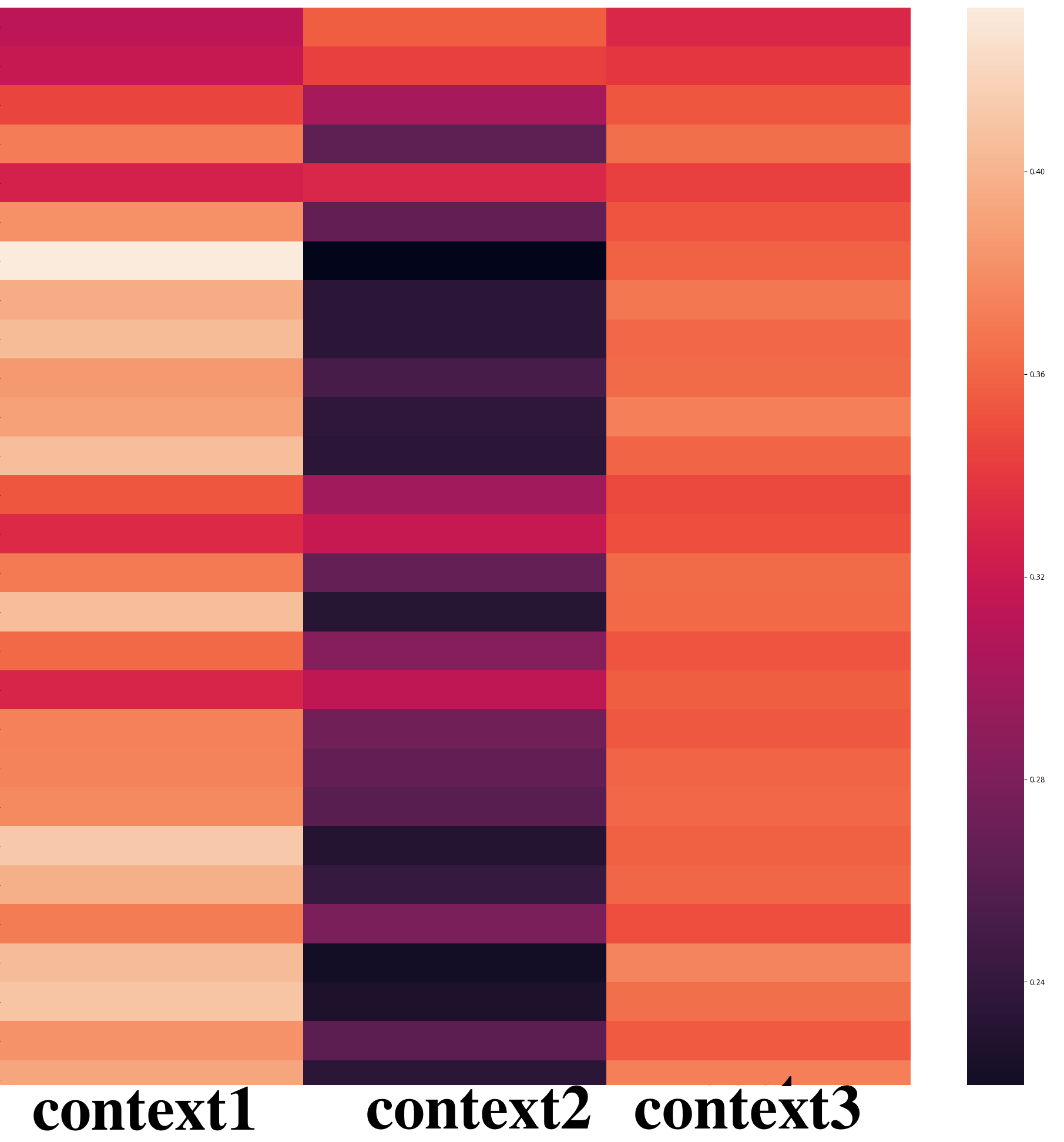}
\end{minipage}
}%

\subfigure[\scriptsize ReCoSa-head5.]{
\begin{minipage}[t]{0.4\linewidth}
\centering
\includegraphics[width=1.4in]{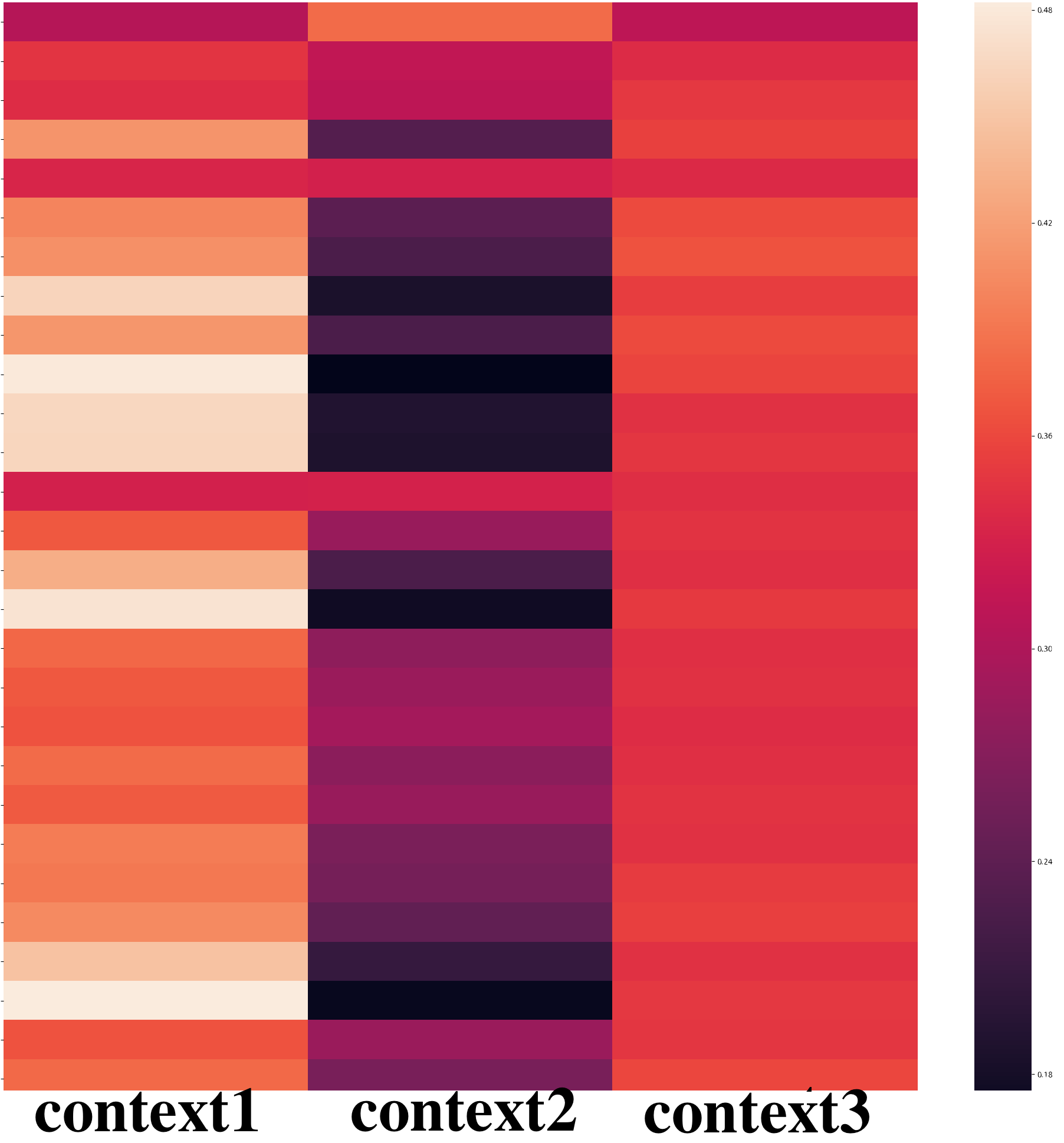}
\end{minipage}
}%
\subfigure[\scriptsize ReCoSa-head6.]{
\begin{minipage}[t]{0.4\linewidth}
\centering
\includegraphics[width=1.4in]{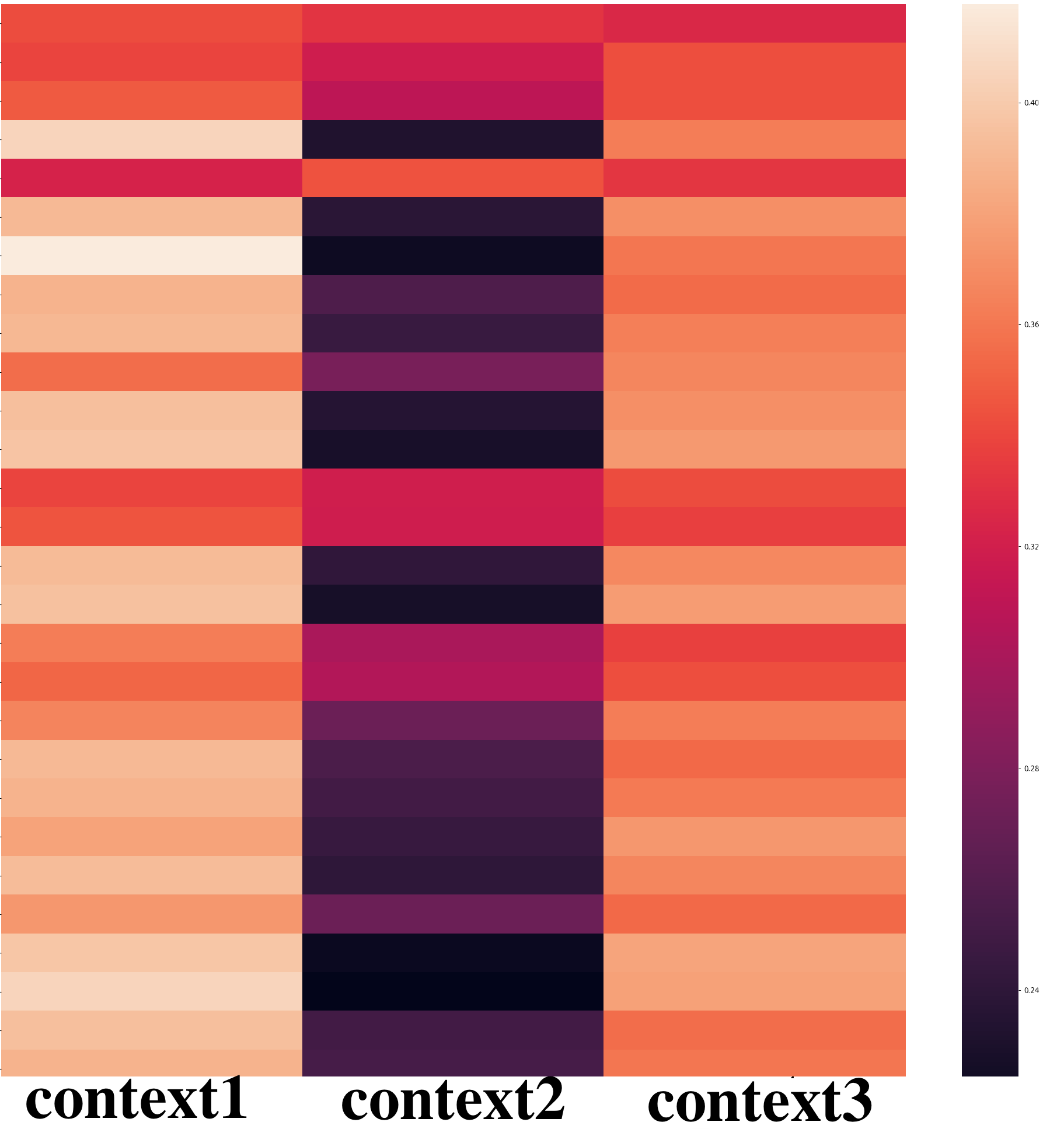}
\end{minipage}
}%
\centering
\caption{\label{fig:caseJDC1}ReCoSa multi-head attention for example1 in Table~\ref{tb:caseJDC1}. The x-coordinate shows the context sentences and the y-coordinate shows the generated words.}
\end{figure}

\subsection{Analysis on Relevant Contexts}
To verify whether the performance improvements are owing to the detected relevant contexts, we conduct a further data analysis, including both quantitative evaluation and case study. Specifically, we randomly sample 500 context-response pairs from the JDC dataset, denoted as JDC-RCD\footnote{https://github.com/zhanghainan/ReCoSa}. Three annotators are employed (all CS PhD students) to label each context with respect to the human\rq{}s judgements. If a contextual sentence is related with the response, then it is labeled as 1. Otherwise it is labeled as 0. The kappa value of this labeled dataset is 0.514, indicting the consistance among different annotators.

\subsubsection{Quantitative Evaluation}
Since HRED considers all the context as relevant context, we calculate the error rate for evaluation. That is, one minus the proportion of all-context-relevant in the JDC-RCD data, i.e.~98.4\%. Therefore, using all contexts indiscriminately is highly inappropriate for multi-turn dialogue generation.

\begin{table*}
\centering
\footnotesize
\newcommand{\tabincell}[2]{\begin{tabular}{@{}#1@{}}#2\end{tabular}}
\begin{tabular}{lll} 
\multicolumn{3}{c}{Example2}\\
\toprule
context1 &  \tabincell{l}{请稍等,正在为您确认此前咨询内容}& \tabincell{l}{Please wait, I am checking the previous requirement.}\\ 
\textcolor{red}{ context2} &  \tabincell{l}{ \textcolor{red}{ 先直接退货再买吗?}}& \tabincell{l}{\textcolor{red}{ Should I return the goods firstly and buy it again?}}\\ 
context3 &  \tabincell{l}{是.收到了吧?}& \tabincell{l}{ Yes. Do you receive the goods?}\\ 
context4 &  \tabincell{l}{对,昨晚收到的}& \tabincell{l}{Yes, I received them last night.}\\ 
{response } &  \tabincell{l}{{ 那就只能无理由退了再买}}& \tabincell{l}{{ You should return them and buy them again}}\\  \hline
Seq2Seq &  \tabincell{l}{好的,亲爱滴~请您稍等下哦}& \tabincell{l}{Ok, dear~Please wait}\\ 
HRED &  \tabincell{l}{可以的}& \tabincell{l}{Ok}\\ 
VHRED &  \tabincell{l}{京东客服很高兴为您服务! }& \tabincell{l}{JD customer services are happy to serve you.}\\ \hline
WSeq &  \tabincell{l}{请问还有其他的问题需要处理呢?}& \tabincell{l}{Do you have any other questions?}\\ 
HRAN &  \tabincell{l}{好.您可以换售后退货了哦}& \tabincell{l}{Ok, you can return the goods.}\\ 
HVMN &  \tabincell{l}{好的,正在查询!回复不够及时请见谅! }& \tabincell{l}{Ok, Searching! Replying to you is not timely enough, sorry!}\\  \hline
{ReCoSa}&  \tabincell{l}{ { 您申请售后,商品退回,重新下单购买}}& \tabincell{l}{You can apply for sale, return the goods and place an order again}\\ 
\bottomrule
\end{tabular}
\caption{\label{tb:caseJDC2}  The generated response Example2 from different models on JDC. The red contexts indicate the relevant context to the response.}
\end{table*}

Other models, such as WSeq, HRAN and HVMN, will output the relevance score based on the attention weight for each context. Therefore we can treat it as a ranking problem. Ranking evaluation measures, such as the precision, recall and F1 score, are used for quantitative evaluations\footnote{To clearly analyze the attention location, we only pay attention to the context without the post. The reason is that the WSeq always rank the post as top1.}. Then we calculate the precision, recall and F1 score of the top 1,3,5,10 for WSeq model, HRAN model and our ReCoSa model.\footnote{The HVMN model doesn\rq{}t have the explicit attention weight, since its memory is based on the variational network.} The results are shown in Table~\ref{tb:ATTENTION}. We can see that the WSeq obtains the best score for P@1, R@1 and F1@1. That\rq{}s because there are 80\% cases that the post is labeled as 1, and the cosine similarity can rank the explicitly similar context sentence as top 1. Though the WSeq has the best score for F1@1, it  doesn\rq{}t work well for F1@3, F1@5 and F1@10. That\rq{}s because the WSeq may lose some relevant contexts which are not explicitly similar to the post but are related with the response. Compared with the HRAN and WSeq, ReCoSa performs better in most cases. Take P@3 for example, the P@3 score of ReCoSa-head3 is 26.2, which is significantly better than that of HRAN and WSeq, i.e.,~24.13 and 24.27. These results indicate that the relevant contexts detected by our ReCoSa model are highly coherent with human\rq{}s judgments. Furthermore, we calculate the averaged attention distance to the response, defined as:
\begin{equation*}
dis2resp = \sum_{i=1}^{N} \frac{N-i+1}{N+1}w_i,
\end{equation*}
where $i$ is the index of the context sentence $s_i$ and $w_i$ is the attention weight of the $i^{th}$ context. The $dis2resp$ in human label is 0.399, indicting that the distribution of human attention is approximately uniform, containing both long and short distant dependencies. The $dis2resp$ in ReCoSa is 0.477, which is closer to human than the distance in HRAN, e.g.~0.291. That is to say, our ReCoSa model can well capture the long distant dependency as compared with traditional attention on HRED, validating the correctness of our ideas.

\begin{figure}[!t]
\centering
\subfigure[\scriptsize ReCoSa-head1.]{
\begin{minipage}[t]{0.4\linewidth}
\centering
\includegraphics[width=1.4in]{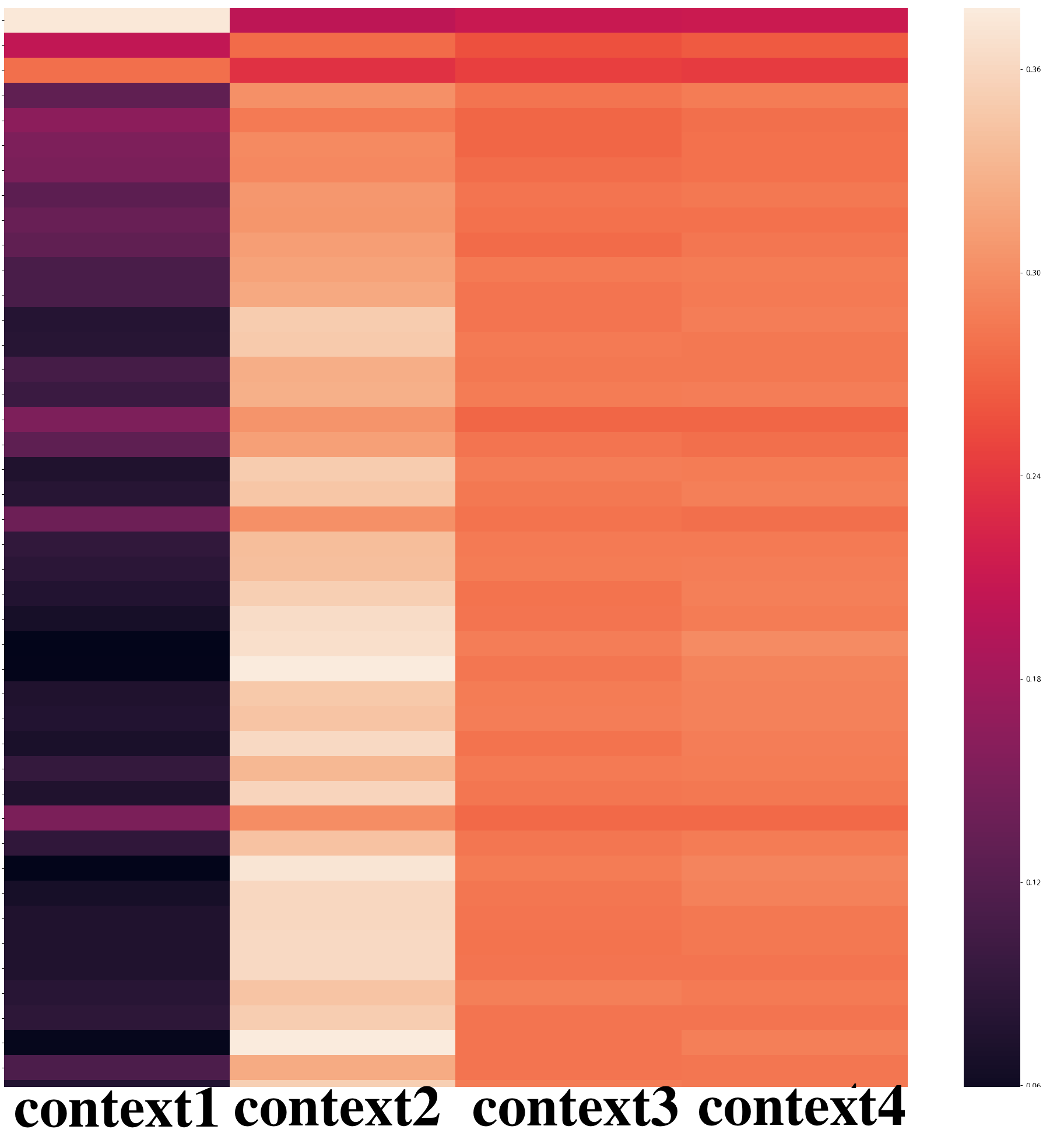}
\end{minipage}%
}%
\subfigure[\scriptsize ReCoSa-head2.]{
\begin{minipage}[t]{0.4\linewidth}
\centering
\includegraphics[width=1.4in]{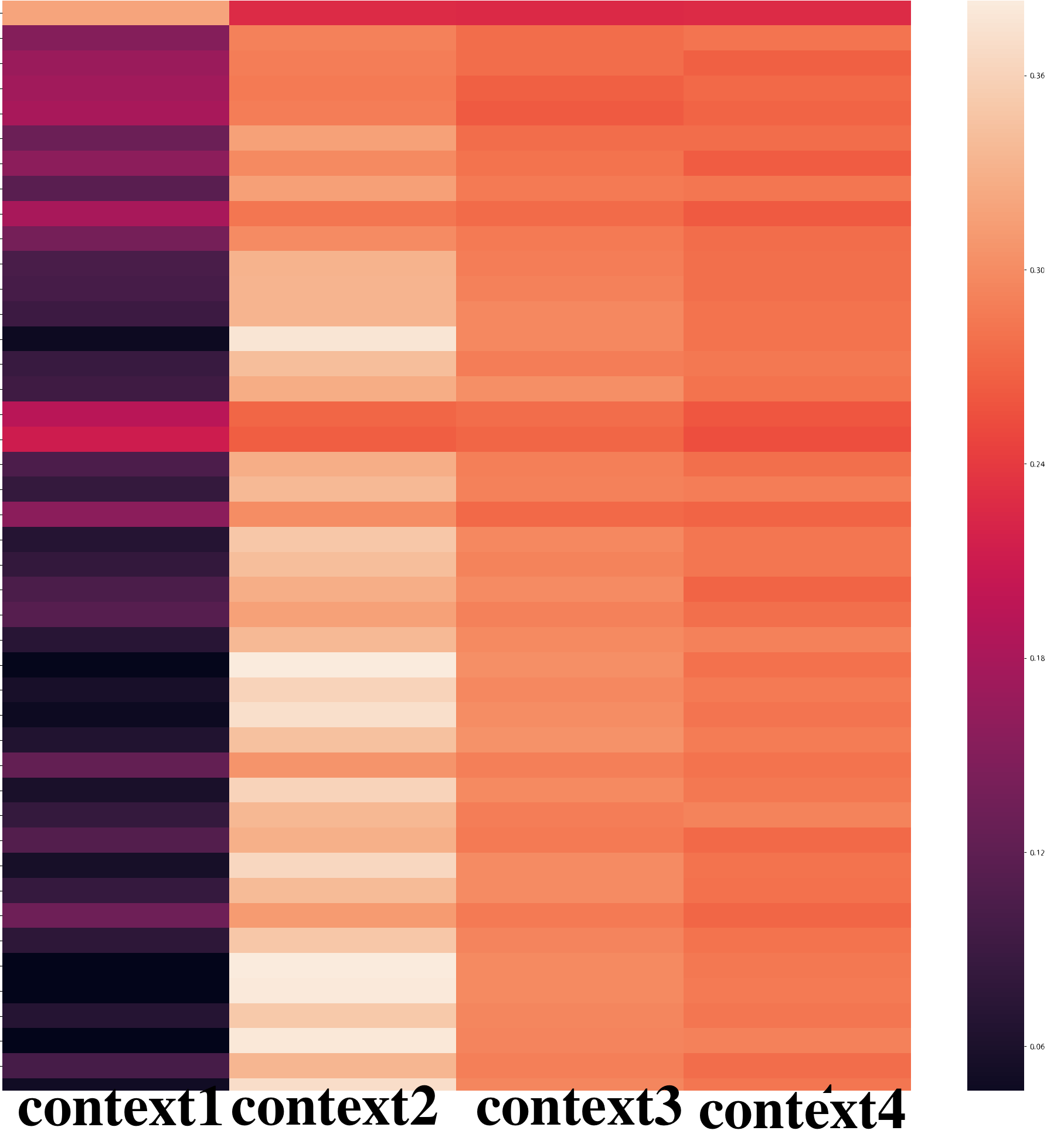}
\end{minipage}%
}%

\subfigure[\scriptsize ReCoSa-head3.]{
\begin{minipage}[t]{0.4\linewidth}
\centering
\includegraphics[width=1.4in]{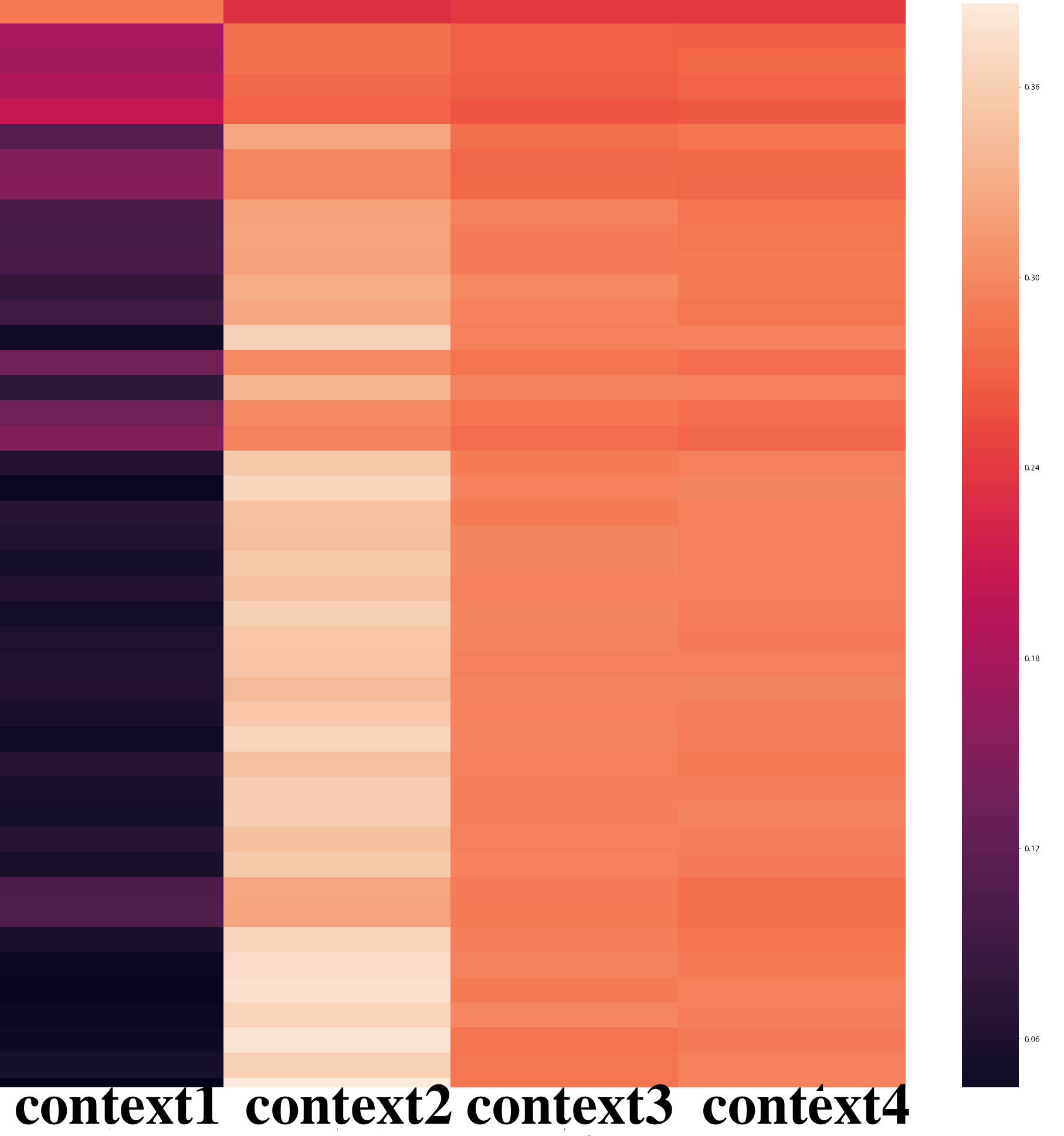}
\end{minipage}
}%
\subfigure[\scriptsize ReCoSa-head4.]{
\begin{minipage}[t]{0.4\linewidth}
\centering
\includegraphics[width=1.4in]{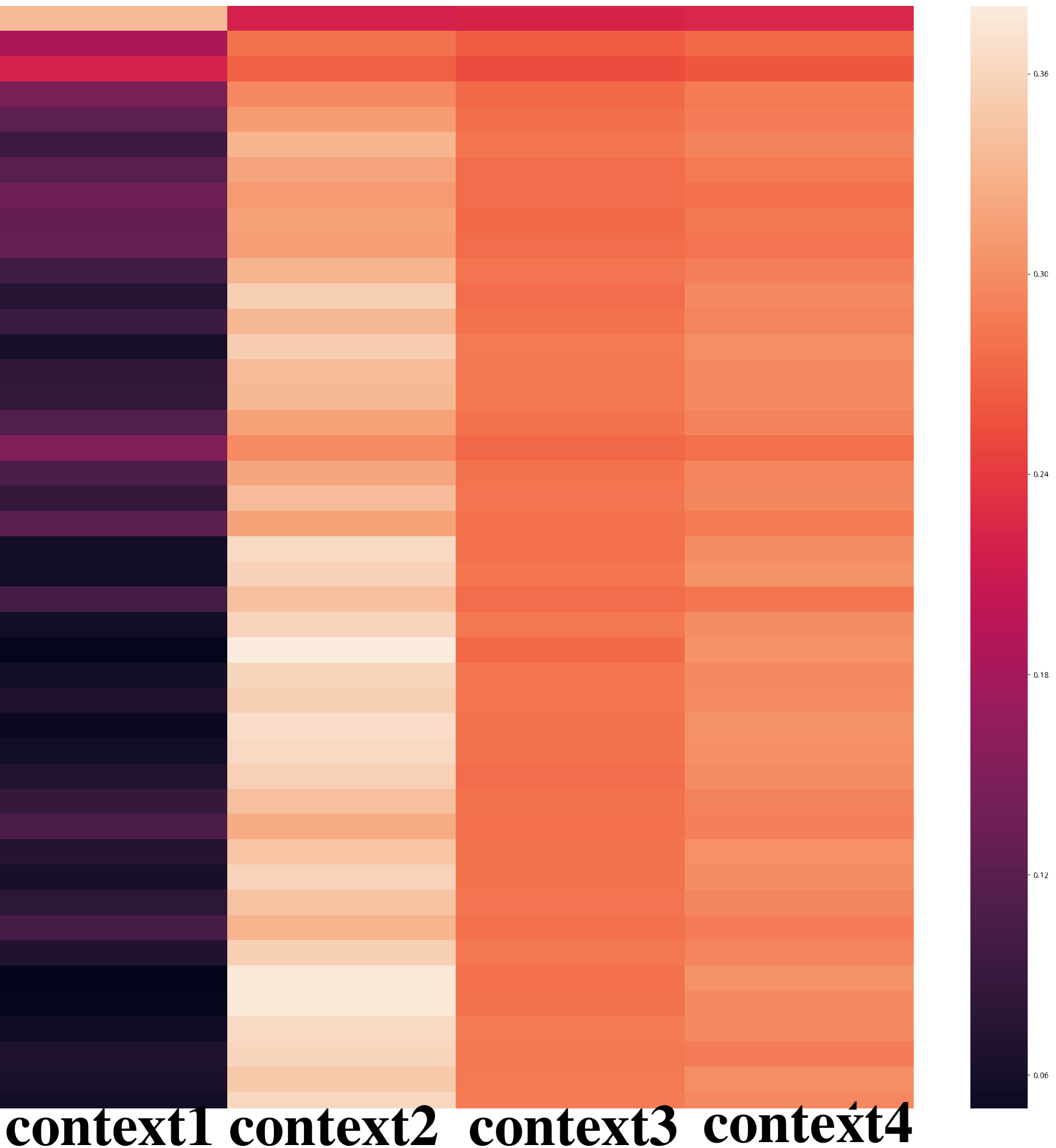}
\end{minipage}
}%

\subfigure[\scriptsize ReCoSa-head5.]{
\begin{minipage}[t]{0.4\linewidth}
\centering
\includegraphics[width=1.4in]{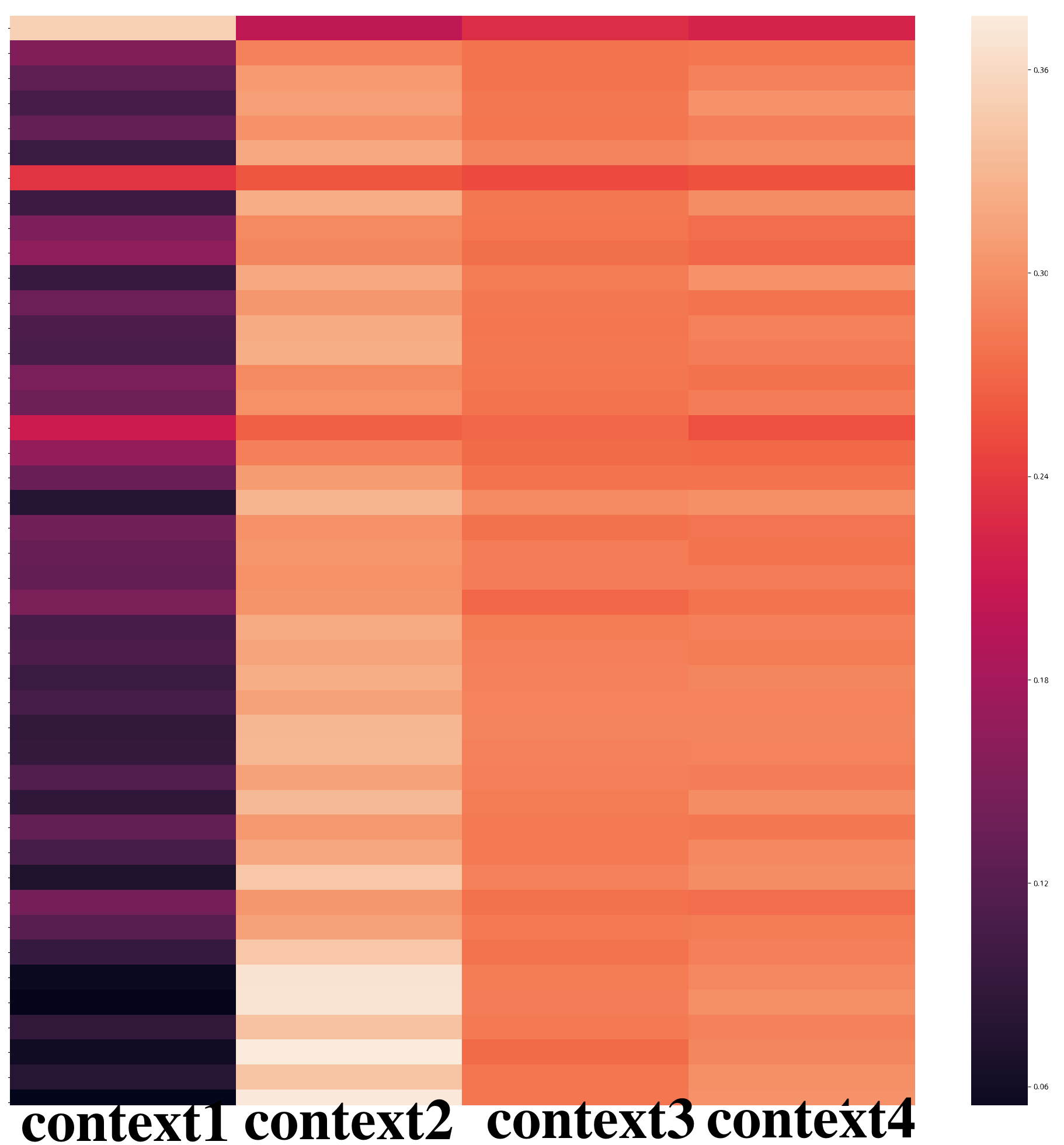}
\end{minipage}
}%
\subfigure[\scriptsize ReCoSa-head6.]{
\begin{minipage}[t]{0.4\linewidth}
\centering
\includegraphics[width=1.4in]{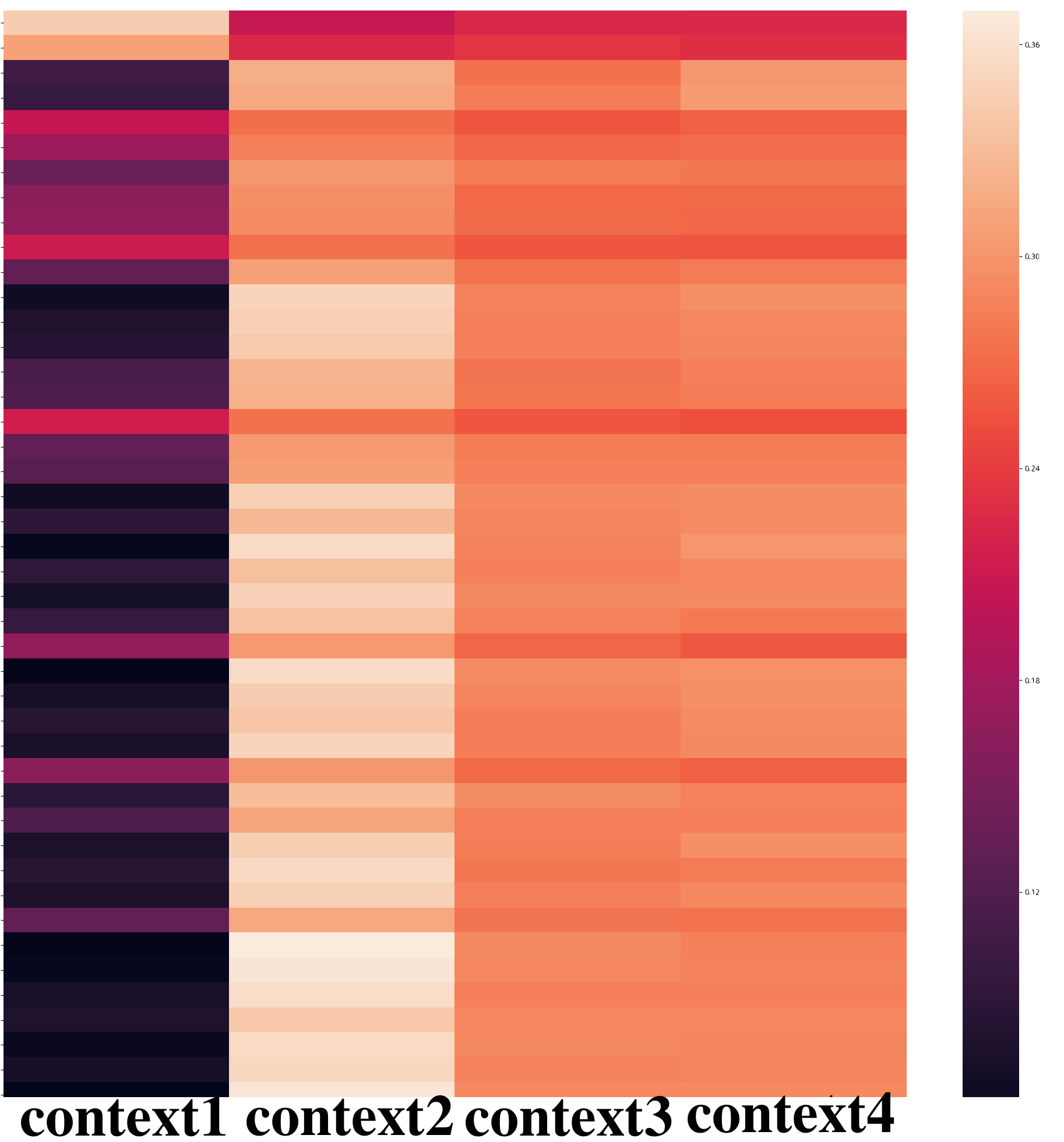}
\end{minipage}
}%
\centering
\caption{\label{fig:caseJDC2}ReCoSa multi-head attention for example2 in Table~\ref{tb:caseJDC2}. The x-coordinate shows the context sentences and the y-coordinate shows the generated words.}
\end{figure}

\subsubsection{Case Study}
To facilitate a better understanding of our model, we give some cases as in Table~\ref{tb:caseJDC1} and~\ref{tb:caseJDC2}, and draw the heatmap of our ReCoSa model, including the six heads, to analyze the attention weights in Figure~\ref{fig:caseJDC1} and~\ref{fig:caseJDC2}. 

From the result, we can first see that the attention-based model performs better than the model using all contexts indiscriminately. Take example1 of Table~\ref{tb:caseJDC1} as an example. The baselines of using all contexts are easy to generate some common responses, such as \lq{}{\em What can I do for you?}\rq{} and \lq{}{\em I am very happy to serve you.} \rq{}. The attention-based models, i.e. HRAN, WSeq, ReCoSa, can generate relevant response, such as \lq{}{\em Applying}\rq{} and \lq{}{\em It\rq{} s already done, and the system can\rq{} t intercept the site.}\rq{}. The response generated by our ReCoSa is more specific and relevant, i.e. \lq{}{\em Your servers order has not been updated yet, please wait.}\rq{}. The reason is that ReCoSA considers the difference of contexts and it will focus on the relevant contexts, i.e.~context1 and context3. Figure~\ref{fig:caseJDC1} shows the heatmap of example1 in Table~\ref{tb:caseJDC1}. The x-coordinate indicates the context1, context2 and context3. And the y-coordinate indicates the generated words. The lighter the color is, the larger the attention weight is. We can see that the ReCoSa pays more attention to the relevant contexts, i.e.~context1 and context3, which is coherent with the human\rq{}s understanding. 

Our model also performs well in the case where the post (i.e.~the closest context) and the ground-truth response are not in the same topic. From the example2 in Table~\ref{tb:caseJDC2},
the baselines all produce irrelevant or common responses, such as {\em  \lq{}Do you have any other questions?\rq{}} and {\em  \lq{}Ok, I am looking for you!  Replying to you is not timely enough, sorry!\rq{}}. The reason is that the baseline models are weak in detecting long distant dependency relations. However, our model gives more relevant responses with specific meanings{\em \lq{}You could apply for sale, return the goods and place an order again\rq{}}, by using the self-attention mechanism. Figure~\ref{fig:caseJDC2} shows the heatmap of example2 in Table~\ref{tb:caseJDC2}. For example2, the context2 is the most significant context and the context1 is the most useless one. We can see that the ReCoSa ignores the context1 and pays more attention to the context2. In a word, our ReCoSa model can detect both the long and short distant dependencies, even for the difficult case when the response is not related with the post. 

\section{Conclusion}
In this paper, we propose a new multi-turn dialogue generation model, namely ReCoSa. The motivation comes from the fact that the widely used HRED based models simply treat all contexts indiscriminately, which violate the important characteristic of multi-turn dialogue generation, i.e., the response is usually related to only a few contexts. Though some researchers have considered using the similarity measure such as cosine or traditional attention mechanism to tackle this problem, the detected relevant contexts are not accurate, due to either insufficient relevance assumption or position bias problem. Our core idea is to utilize the self-attention mechanism to effectively capture the long distant dependency relations. We conduct extensive experiments on both Chinese customer services dataset and English Ubuntu dialogue dataset. The experimental results show that our model significantly outperforms existing HRED models and its attention variants. Furthermore, our further analysis show that the relevant contexts detected by our model are significantly coherent with humans\rq{} judgements. Therefore, we obtain the conclusion that the relevant contexts can be useful for improving the quality of multi-turn dialogue generation, by using proper detection methods, such as self-attention. 

In future work, we plan to further investigate the proposed ReCoSa model. For example, some topical information can be introduced to make the detected relevant contexts more accurate. In addition, the detailed content information can be considered in the relevant contexts to further improve the quality of generated response.
\end{CJK*}
\section*{Acknowledgments}
This work was funded by the National Natural Science Foundation of China (NSFC) under Grants No. 61773362, 61425016, 61472401, 61722211, and 61872338, the Youth
Innovation Promotion Association CAS under Grants No.20144310, and 2016102, and the National Key R\&D Program of China under Grants No. 2016QY02D0405.
\bibliographystyle{acl_natbib}|
\bibliography{acl2019}| 
\end{document}